\documentclass[journal]{IEEEtran}

\usepackage{amsmath}
\usepackage{amssymb}
\usepackage{booktabs}
\usepackage{array}
\usepackage{tabularx}
\usepackage{graphicx}
\usepackage{stfloats}
\usepackage{verbatim}
\usepackage{url}
\usepackage{cite}
\usepackage[table]{xcolor}
\usepackage{tikz}
\usetikzlibrary{arrows.meta,positioning,fit,calc,shapes.geometric}
\usepackage{hyperref}

\hypersetup{
  colorlinks=true,
  linkcolor=blue,
  citecolor=blue,
  urlcolor=blue
}

\newcommand{\best}[1]{\textbf{#1}}
\newcommand{\second}[1]{\underline{#1}}

\title{Steins;Gate Drive: Semantic Safety Arbitration over Structured Futures for Latency-Decoupled LLM Planning}
\author{Anjie~Qiu and Hans~D.~Schotten%
\thanks{Anjie Qiu and Hans D. Schotten are with the Institute for Wireless Communication and Navigation, RPTU University Kaiserslautern-Landau, Kaiserslautern, Germany (e-mail: \{aqiu, schotten\}@rptu.de).}%
\thanks{Hans D. Schotten is also with the Intelligent Networks Research Group (IN), German Research Center for Artificial Intelligence (DFKI), Kaiserslautern, Germany (e-mail: schotten@dfki.de). Corresponding author: Hans D. Schotten.}}

\begin{document}

\markboth{IEEE Transactions Manuscript,~2026}%
{Qiu and Schotten: Steins;Gate Drive}

\maketitle

\begin{abstract}
Cloud-hosted LLM driver agents provide useful semantic judgments, but their inference latency exceeds stepwise vehicle-control windows. Learned world models predict futures, but they usually keep future generation and action selection inside large coupled loops. We present \textbf{Steins;Gate Drive}, a latency-decoupled planner-runtime architecture in which the world-line metaphor from the eponymous story names one plausible consequence of an intervention: the LLM selects counterfactual driving futures before the final control instant, and a runtime reuses the selected forecast only while safety contracts remain valid. The generator builds three world-line roles: \texttt{alpha} nominal ego-conditioned futures, \texttt{beta} interaction counterfactuals around nearby vehicles, and \texttt{gamma} hazard-stress futures such as braking, cut-ins, or blocked corridors. The selected branch becomes a typed \texttt{StrategicForecast} with horizon, validity/abort conditions, fallback, and authority. On a within-subject, matched-seed normal-highway protocol with 10 seeds and 20 steps, GPT-5.4 mini reduces effective lag from $+3.07$~s at 1-second horizon to $-0.01$~s at 4-second horizon while preserving the measured no-collision safety boundary. The architecture's safety contribution comes from the atom-predicate runtime check, not from the drift score, which functions as a refresh-frequency knob.
\end{abstract}


\section{Introduction}

\IEEEPARstart{A}{utonomous} driving exposes a sharp timing problem for large language models. A closed-loop agent must issue actions while traffic is still changing, but the semantic reasoning that makes LLMs attractive can take longer than a control step. If the model is asked to choose only at the last reactive instant, its answer may arrive after the scene has already changed; if the model is allowed to plan earlier, the system must still know when that plan has expired.

We study that interface problem. Rather than making the LLM faster or replacing the low-level driving loop with free-form language, the method makes a slow semantic choice usable by turning it into a reusable object with explicit validity conditions, abort conditions, fallback behavior, and runtime authority. This motivates the question: \textbf{under what architecture can a slow LLM's semantic safety reasoning matter before the control window closes?}

Prior work already moves LLM driving away from direct fast control. LLM4AD~\cite{cui2024llm4ad} identifies latency, deployment, and trust as core barriers. CoT-Drive~\cite{liao2025cotdrive}, LeAD~\cite{zhang2025lead}, DiMA~\cite{hegde2025distilling}, and world-model/VLA systems~\cite{feng2025worldmodelsad} move toward forecasting, lower-frequency reasoning, distillation, or anticipatory planning. The missing piece is the planner-facing interface: how can a stronger LLM remain the final semantic decision-maker when its raw inference time is too slow for reactive driving?

\begin{table}[!t]
\centering
\caption{Horizon-sweep teaser: increasing the forecast-reuse horizon amortizes effective lag while preserving no-collision rate and exposing driving-behavior trade-offs. Values are from the GPT-5.4 mini dual-agent realistic-highway \texttt{n10}/20-step sweep.}
\label{tab:horizon_resolved}
{\scriptsize
\setlength{\tabcolsep}{2.2pt}
\renewcommand{\arraystretch}{1.04}
\resizebox{\columnwidth}{!}{%
\begin{tabular}{@{}lccccc@{}}
\toprule
Metric & $H{=}1$ s & $H{=}2$ s & $H{=}4$ s & $H{=}8$ s & $H{=}12$ s \\
\midrule
No-col. rate \%$\uparrow$ & 100.0 & 100.0 & 100.0 & 100.0 & 100.0 \\
Speed km/h$\uparrow$ & \best{73.4} & \second{70.2} & 68.7 & 67.1 & 50.0 \\
Flap \%$\downarrow$ & \best{2.5} & \second{5.0} & 11.0 & 7.0 & \best{2.5} \\
Eff. lag s$\downarrow$ & +3.07 & +1.97 & -0.01 & \second{-3.98} & \best{-7.79} \\
\bottomrule
\end{tabular}}
\vspace{1pt}
\begin{minipage}{\columnwidth}
\scriptsize
\emph{Note.} Speed is mean ego speed; flap is the action-switching rate. Eff. lag follows (\ref{eq:effective_lag}); negative values mean the buffered horizon covers more time than the measured call cost. For a 10/10 no-collision cell, the Wilson 95\% interval is $[72.2,100.0]$; full intervals are in the supplementary material. Bold marks the best comparable value and underlining marks the second-best value for directed rows; fully tied rows are left unranked.
\end{minipage}
}
\end{table}

We propose \textbf{Steins;Gate Drive}, a latency-decoupled planner-runtime architecture whose design principle is \textbf{structured, early, and revocable} LLM reasoning. At strategic time $t$, the system extracts a compact symbolic state $S_t=(E_t,V_t,M_t)$, generates and prunes \texttt{alpha}/\texttt{beta}/\texttt{gamma} future branches, and asks the strategic LLM to select one branch. The output is a typed \texttt{StrategicForecast}, not an unconditional action: it carries branch choice, validity atoms, abort atoms, fallback action, horizon, provenance, and authority level, and the canonical dual-agent runtime executes, softens, overrides, or refreshes that forecast as the live scene evolves. This framing differs from predictor-centric systems such as CoT-Drive, lower-frequency semantic controllers such as LeAD, and unified latent planners such as DriveWorld-VLA~\cite{jia2026driveworldvla}: Steins;Gate Drive uses forecasting only to expose a compact choice set, keeps a strong LLM as the branch selector, and makes the selected future reusable only while machine-checkable assumptions remain valid. The comparison is architectural positioning rather than an empirical superiority claim; the experiments below compare only implemented modes under matched highway seeds. \S\ref{sec:method} formalizes the architecture; \S\ref{sec:env_setup}--\S\ref{sec:results} evaluate it. This paper makes the following four contributions, each tested empirically in \S\ref{sec:results}:

\begin{enumerate}
\item We introduce a role-typed world-line generator that converts each feasible primitive highway action into nominal \texttt{alpha}, interaction-counterfactual \texttt{beta}, and hazard-stress \texttt{gamma} futures.
\item We introduce a two-timescale buffered planner-runtime architecture in which the LLM emits a typed, revocable \texttt{StrategicForecast} rather than a last-instant action.
\item We evaluate the forecast-buffer architecture against reactive and deterministic-replay baselines on matched $n{=}10$/20-step normal highway runs, framing the comparison as a deployability diagnostic rather than a safety-superiority claim.
\item We identify the horizon sweep as the main latency-amortization result: GPT-5.4 mini moves from positive to near-zero effective lag while role-set, branch-budget, and drift-threshold sweeps diagnose the behavior trade-offs behind that result.
\end{enumerate}

\section{Related Work}

\subsection{Language and End-to-End Driving}

Three lineages of language-based driving directly precede this work: direct LLM/VLM drivers, lower-frequency LLM planners, and learned end-to-end driving stacks. GPT-Driver~\cite{mao2023gptdriver}, DriveGPT4~\cite{xu2024drivegpt4}, DiLu~\cite{wen2024dilu}, and LLM4AD~\cite{cui2024llm4ad} motivate semantic driving decisions while exposing latency, deployment, and trust barriers. CoT-Drive~\cite{liao2025cotdrive}, LeAD~\cite{zhang2025lead}, and DiMA~\cite{hegde2025distilling} move high-level reasoning away from raw high-frequency control through forecasting, lower-frequency reasoning, or distillation. End-to-end planners such as UniAD~\cite{hu2023uniad} and VAD~\cite{jiang2023vad} optimize learned perception-prediction-planning stacks. Steins;Gate Drive contributes at this layer a cacheable and revocable external-LLM decision interface, not another direct LLM driver or learned perception stack.

\subsection{World Models and Future Prediction}

World-model work addresses a complementary problem: the system must reason about futures, not only the current frame. One branch generates future scenes or occupancy states, including GAIA-1~\cite{hu2023gaia1}, Copilot4D~\cite{zhang2023copilot4d}, DriveDreamer~\cite{wang2023drivedreamer}, DriveDreamer-2~\cite{zhao2024drivedreamer2}, GenAD~\cite{zheng2024genad}, and Driving in the Occupancy World~\cite{yang2024driveoccworld}. A second branch learns latent imagination or latent planning, from World Models~\cite{ha2018worldmodels} and Dreamer~\cite{hafner2020dreamer} to LAW~\cite{li2025law} and World4Drive~\cite{zheng2025world4drive}. Unified VLA-style stacks such as DriveWorld-VLA~\cite{jia2026driveworldvla} extend this direction toward joint world modeling and action choice. Steins;Gate Drive contributes bounded \texttt{alpha}/\texttt{beta}/\texttt{gamma} future construction as a planner-facing interface, not a learned visual simulator or Dreamer-style behavior model.

\subsection{Revocable Reasoning-Action Interfaces}

Revocable action interfaces are needed whenever a useful future choice may become stale before execution. Model predictive control~\cite{camacho2007modelpredictive} provides the classical template for future-conditioned action under repeated state updates, and safety frameworks such as RSS~\cite{shalev2017rss} and TTC indicators~\cite{vogel2003ttc} separate runtime checks from nominal planning. Latent-world-model planners such as LAW and World4Drive blur the line, but their revocation mechanism is implicit in latent replanning or retraining rather than explicit atom predicates. Steins;Gate Drive contributes a machine-checkable LLM forecast contract, not an optimized continuous control sequence: a strong LLM remains an external selector over explicit world lines, while validity, abort, fallback, and authority fields bind the selected future to runtime execution.

\section{Method: Role-Typed World Lines and Forecast-Buffered Runtime}
\label{sec:method}

\subsection{Overview and Design Thesis}

Steins;Gate Drive treats LLM driving as a two-timescale decision problem. At control step $t$, the vehicle observes a live highway scene $x_t$ and must issue a discrete action before the scene evolves, but a strong cloud-hosted LLM may require longer than one control period to compare alternatives. The method does not ask the LLM for a last-instant action. It constructs a bounded set of counterfactual driving futures, asks the LLM to select one future early, and stores that selection as a revocable forecast contract. If the forecast horizon covers the measured strategic and runtime call cost, the effective lag can approach zero without making the LLM itself faster.

\begin{figure*}[t]
\centering
\includegraphics[width=\textwidth]{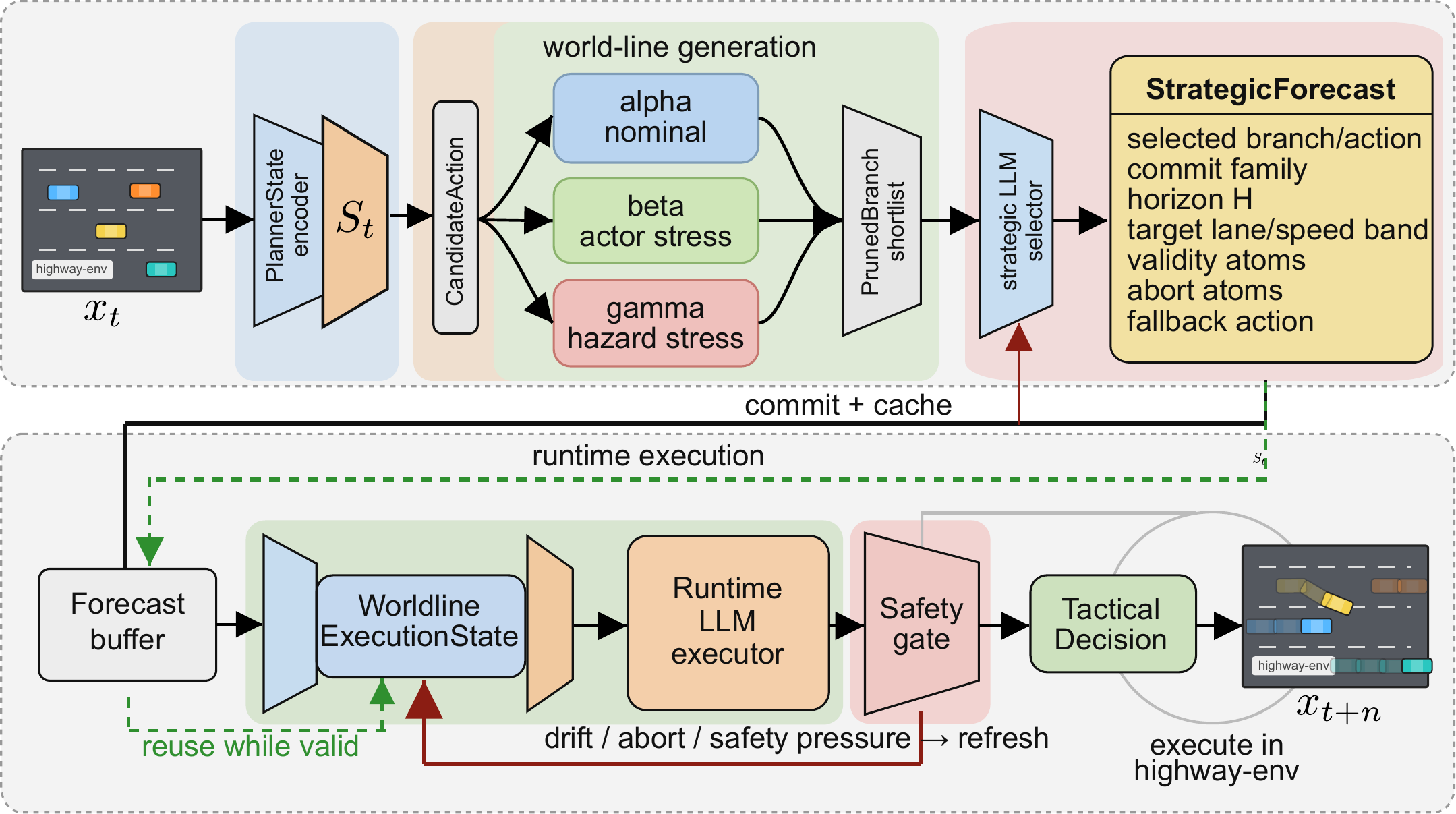}
\caption{Overall Steins;Gate Drive architecture. A compact highway state and feasible primitive actions define role-typed world-line branches. Analytical scoring produces a short selector-visible set; the strategic LLM commits one branch as a typed \texttt{StrategicForecast}. The canonical dual-agent runtime stores this object in a forecast buffer, tracks \texttt{WorldlineExecutionState}, and emits a constrained \texttt{TacticalDecision}; validity, abort, drift, or safety pressure triggers override or strategic refresh.}
\label{fig:contract_flow}
\end{figure*}

The core object is the role-typed world line. A world line is an action-conditioned future over a short horizon $H$ that exposes the consequence of one ego maneuver, one stressed neighbor interaction, or one configured hazard. It is not a learned visual world model: no pixel-level future scene is generated and no hidden latent planner is assumed. The method follows the module flow in Fig.~\ref{fig:contract_flow}: encode a compact highway state, generate and score role-typed world lines, ask the LLM to commit one as a typed forecast, then reuse that forecast only while runtime checks permit it. The following subsections define the state/action interface, generator, contract, and runtime; the corresponding mechanism effects are reported in \S\ref{sec:results}.

\subsection{World-Line State Encoding}

The strategic planner receives a compact state
\begin{equation}
S_t=(E_t,V_t,M_t),
\end{equation}
where $E_t$ summarizes the ego vehicle, $V_t=\{v_i\}$ contains the nearest retained traffic participants, and $M_t$ stores lane/action metadata and fallback information. This representation is extracted from the live \texttt{highway-env}~\cite{leurent2018highwayenv} scene and is the only scene input used for branch construction. It removes image-level detail while preserving the quantities needed for highway decision making: lane position, speed, relative gaps, relative speed, and action feasibility.

The action set $\mathcal{A}_t$ is the feasible subset of the five discrete environment actions: \texttt{LANE\_LEFT}, \texttt{IDLE}, \texttt{LANE\_RIGHT}, \texttt{FASTER}, and \texttt{SLOWER}. A \texttt{CandidateAction} keeps the primitive action ID, the action token, the target lane when applicable, and a constant-acceleration proxy used only for short-horizon rollout. Primitive actions remain the simulator boundary; higher-level maneuver meaning is introduced later through the forecast commit family.

\begin{figure*}[t]
\centering
\includegraphics[width=\textwidth]{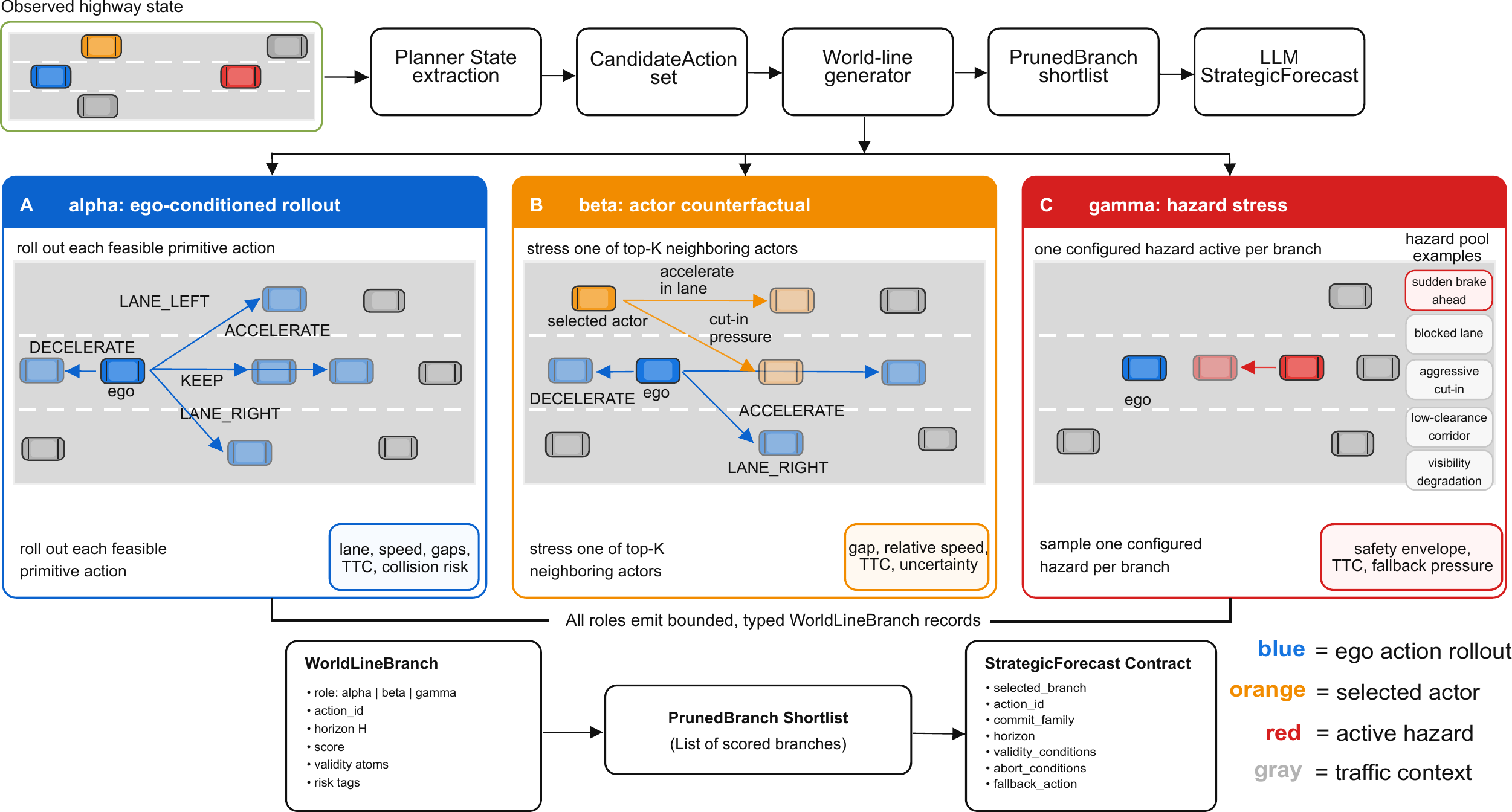}
\caption{World-line generation roles in a \texttt{highway-env}-style highway scene. Alpha branches roll out each feasible primitive action with ego kinematics and local gaps. Beta branches reuse the same action-conditioned future but stress a selected neighboring actor. Gamma branches reuse the action-conditioned future but apply a configured hazard. The generator produces a bounded, typed branch set in which every future has a reproducible source and paper-facing role tag.}
\label{fig:worldline_generation_roles}
\end{figure*}

\subsection{Role-Typed World-Line Generation}
\label{sec:worldline_generation}

\paragraph{Generator interface.} The generator $\Phi_{\alpha\beta\gamma}$ takes $(S_t,\mathcal{A}_t,H)$ and returns a finite branch set
\begin{equation}
\mathcal{B}_t=\mathcal{B}^{\alpha}_t \cup \mathcal{B}^{\beta}_t \cup \mathcal{B}^{\gamma}_t .
\end{equation}
Each branch $b\in\mathcal{B}_t$ stores the primitive action, predicted lane/speed/progress summary, front and rear gaps, minimum time-to-collision (TTC), collision risk, comfort and uncertainty penalties, role tags, and a short language summary for the selector. Numerical constants, perturbation offsets, and budget knobs are reported in the supplementary material.

\paragraph{Role semantics.} The three roles separate different sources of future uncertainty and draw on action-conditioned rollouts in model predictive control, critical-actor counterfactuals in interactive driving, and configured safety perturbations. Compared with visual future generators such as GAIA-1, DriveDreamer, and DriveDreamer-2, the trade-off is explicitness and bounded cost rather than open-ended scene synthesis.

\paragraph{Alpha (ego rollout).} For each $a \in \mathcal{A}_t$, the generator resolves the target lane (the requested adjacent lane for lane changes; the current lane otherwise) and rolls out constant-acceleration kinematics over $H$: $\hat v_a = \max(0, v_{\mathrm{ego}} + u_a H)$ and $\hat p_a = \max(0, v_{\mathrm{ego}} H + \tfrac{1}{2} u_a H^2)$, with $u_a$ the action acceleration proxy. Front and rear gaps are measured to the closest target-lane vehicles. The adverse alpha variant retains the primitive action but applies bounded gap/TTC reductions before scoring risk via the proxy
\begin{equation}
r = \min\!\Big\{1,\; \textstyle\sum_{k\in\{T,f,r\}} w_k\,[\tau_k - x_k]_+\,/\,\tau_k\Big\},
\label{eq:collision_risk}
\end{equation}
where $x_T{=}T_{\min}$, $x_f{=}g_f$, $x_r{=}g_r$ and $[x]_+ = \max(x,0)$, consistent with extended-TTC indicators in traffic safety analysis. Weights $(w_k,\tau_k)$ and adverse offsets are listed in the supplementary material.

\paragraph{Beta (actor counterfactual).} Given $V_t = \{v_i\}$, a deterministic critical-actor score
\begin{equation}
\kappa(v_i, a) = \mathbf{1}[\text{role}(v_i, a) \neq \emptyset] \cdot \big(\beta_1 |\Delta v_i| + \beta_2 / \max(g_i, 1)\big)
\label{eq:critical_actor}
\end{equation}
ranks neighbors conditioned on $a$, where $\text{role}(v_i, a)$ is nonempty only when the actor is geometrically relevant ($\Delta v_i$ relative speed, $g_i$ gap). Same-lane front/rear vehicles are eligible for keep/accelerate/decelerate, target-lane vehicles for lane changes, and remaining close vehicles as adjacent actors; the top-$K$ scored neighbors define $\mathcal{B}^{\beta}_t(a)$. Each selected actor perturbs the nominal alpha future with a role-specific stress---\texttt{front\_lead} brake-hard, \texttt{rear\_vehicle} rush-from-behind, \texttt{adjacent\_vehicle} cut-in---while preserving the primitive action and recording the responsible actor ID.

\paragraph{Gamma (hazard stress).} For each feasible action, the generator samples one hazard from a closed pool $\mathcal{H}$ containing grounded highway hazards (sudden front braking, rear aggression, adjacent cut-in, blocked lane, low-clearance corridor) and abstract stressors (unexpected obstacle, visibility degradation). Each hazard applies a fixed penalty tuple to gap, TTC, risk, comfort, and uncertainty, yielding a parameterized stress perturbation while keeping the primitive action and attaching \texttt{hazard\_tags} to expose the stress source to the selector.

\paragraph{Branch-set cardinality.} By construction, the generator emits a finite branch set with bounded cardinality:
\begin{equation}
|\mathcal{B}_t| \le |\mathcal{A}_t| \cdot \big(n_{\alpha} + K + 1\big),
\label{eq:branch_cardinality}
\end{equation}
where $n_{\alpha}$ is the number of alpha variants per action, $K$ is the critical-actor budget for beta, and the ``$+1$'' accounts for one gamma branch per action. This bound governs selector context size and is the world-line budget studied in \S\ref{sec:results}.

\paragraph{Scope boundary.} The role taxonomy gives every future a reproducible source, but it does not provide a learned generative model of the scene. Alpha rollouts use short-horizon kinematics, beta branches stress one critical actor at a time, and gamma branches draw from a closed hazard pool. These limits match the present highway study and are revisited in \S\ref{sec:limitations}.

\paragraph{Analytical scoring and shortlisting.} Every generated branch is converted into a \texttt{PrunedBranch} by a fixed analytical score used as a reproducible pruning reference and as the score-top reference for the lower-score-selection diagnostic. The aggregate score combines clipped TTC and front-gap quality with the collision-risk proxy of (\ref{eq:collision_risk}) and a comfort/uncertainty correction:
\begin{equation}
\begin{aligned}
q_{\mathrm{saf}} =\; &
[w^q_T\min(T_{\min}/5,1) + w^q_g\min(g_f/35,1) - w^q_r r]_+,\\
s_{\mathrm{agg}} =\; &
\lambda_{\mathrm{saf}} q_{\mathrm{saf}}
 + \lambda_{\mathrm{eff}}s_{\mathrm{eff}} \\
& + \lambda_{\mathrm{cmf}}\max\{0,1-c\}
 - \lambda_{\mathrm{unc}}s_{\mathrm{unc}} .
\end{aligned}
\end{equation}
where $c$ is the comfort penalty and $s_{\mathrm{eff}},s_{\mathrm{unc}}$ are efficiency and uncertainty scores; weight vectors are pinned in the supplementary material. Branches are sorted by aggregate, safety, action ID, and branch ID, and the shortlist preserves role diversity before backfilling by score.

\subsection{\texttt{StrategicForecast} Contract}
\label{sec:strategic_forecast}

\paragraph{Typed object.} The selected branch becomes a \texttt{StrategicForecast}, a structured decision object
\begin{equation}
F = \big(b,\; a,\; \phi,\; h,\; V,\; A,\; f,\; \rho,\; \pi,\; t_{\mathrm{issue}}\big),
\label{eq:strategic_forecast}
\end{equation}
where $b \in \mathcal{B}_t$ is the selected branch, $a \in \mathcal{A}_t$ its primitive action, $\phi \in \mathcal{F}$ a commit family, $h \in \mathbb{N}$ a horizon in steps, $V$ a finite set of validity atoms, $A$ a finite set of abort atoms, $f \in \mathcal{A}_t$ a fallback action, $\rho \in \{\mathtt{low}, \mathtt{med}, \mathtt{high}\}$ an authority level, $\pi$ a provenance record, and $t_{\mathrm{issue}}$ the issue step. The forecast is cacheable only if every field has the declared type and every condition atom parses against (\ref{eq:atom_grammar}); otherwise the runtime discards it and falls back to the analytical-top action. This typed-object discipline lets the runtime check applicability without reinterpreting free-form reasoning.

\paragraph{Two-layer action interface.} The action interface is two-layered. Primitive environment actions ($\mathcal{A}_t$) remain the simulator boundary; commit families ($\mathcal{F}$) express what a selected future means over the forecast horizon (e.g., \texttt{KEEP}, \texttt{ACCELERATE}, \texttt{CHANGE\_LEFT}, \texttt{CHANGE\_RIGHT}, \texttt{DECELERATE}). The commit family is what the runtime preserves under authority $\rho$ across multiple steps; the primitive action is what the simulator consumes at any single step.

\paragraph{Atom grammar.} Validity and abort atoms are drawn from a regular grammar over the supported condition vocabulary:
\begin{equation}
\begin{aligned}
\langle\textit{atom}\rangle &::= \langle\textit{metric}\rangle\ \texttt{\_}\ \langle\textit{cmp}\rangle\ \texttt{:}\ \langle\textit{threshold}\rangle\\
\langle\textit{metric}\rangle &::= \mathtt{front\_gap}\ |\ \mathtt{min\_ttc}\ |\ \mathtt{drift\_score}\\
\langle\textit{cmp}\rangle &::= \mathtt{ge}\ |\ \mathtt{gt}\ |\ \mathtt{le}\ |\ \mathtt{lt}\\
\langle\textit{threshold}\rangle &::= \mathbb{R}_{\ge 0}
\end{aligned}
\label{eq:atom_grammar}
\end{equation}
Front-gap thresholds are in meters, TTC thresholds in seconds, and \texttt{drift\_score} thresholds lie in $[0,1]$. Examples are \texttt{front\_gap\_ge:12.0}, \texttt{min\_ttc\_lt:2.0}, and \texttt{drift\_score\_gt:0.35}.

\paragraph{Validity vs.\ abort semantics.} Validity atoms are safety-style invariants that must remain true, while abort atoms are liveness-style stop conditions whose first satisfaction triggers invalidation under (\ref{eq:invalid_predicate}). The rationale field $\pi$ uses a closed tag vocabulary (e.g., \texttt{safety\_margin}); it is not consumed by the runtime decision logic and serves only as a self-reported diagnostic for the offline ledger audit (\S\ref{sec:results}).

\paragraph{Prompt contracts as machine-checked interfaces.} The strategic prompt exposes only a selector-visible shortlist and requires a fixed response template that maps directly to (\ref{eq:strategic_forecast}); the parser accepts only atoms valid under (\ref{eq:atom_grammar}). The runtime prompt is narrower: it consumes the active forecast, live scene, authority level, validity status, and advisory diagnostics, and returns a parseable action line plus an optional one-sentence reason. Invalid forecasts and non-strict runtime outputs are recorded as parser-fallback events.

\paragraph{Role-balanced selector prompt.} The default selector prompt is \emph{natural} (analytical pruning order). In the \emph{balanced} mode used for the role-set sweep in \S\ref{sec:results}, the strategic prompt is augmented with a per-step \texttt{role\_priority} that rotates over the enabled roles in the order $(\alpha,\beta,\gamma)$ as the issued step advances, plus the following fixed instruction block: \emph{``Balanced selection mode: compare alpha, beta, and gamma as equal evidence hypotheses, not as a nominal score ladder. Beta and gamma scores include counterfactual or stress penalties; do not treat those penalties as automatic inferiority to alpha. Use \texttt{role\_priority} only as a tie-breaker when safety, TTC/gap evidence, and robust progress are comparable.''} The balanced block makes the role taxonomy actionable: without it, the LLM defaults to the analytical top branch even when stress branches are visible. Full prompt templates and a representative strategic-then-runtime exchange are in the supplementary material.

\subsection{Forecast-Buffered Runtime Arbitration}
\label{sec:runtime_arbitration}

The runtime layer is a fast supervisor over the slow strategic decision: at every step it decides whether the active forecast remains applicable and falls back to a verified analytical action whenever applicability fails. This places Steins;Gate Drive in the family of two-timescale hierarchical control architectures with an explicit safety supervisor: the LLM is the slow strategic layer, and condition-atom checking is the supervisor.

\paragraph{Drift score.} For a forecast with expected front gap $g_f^*$, expected minimum TTC $T_{\min}^*$, current values $g_f, T_{\min}$, and a lane-mismatch indicator $\ell \in \{0,1\}$, the runtime computes a scalar drift score
\begin{equation}
d = \min\!\left\{1,\; \omega_g\frac{[g_f^*-g_f]_+}{\max(g_f^*,1)}
  + \omega_T\frac{[T_{\min}^*-T_{\min}]_+}{\max(T_{\min}^*,1)}
  + \omega_{\ell}\ell\right\},
\label{eq:drift_score}
\end{equation}
combining front-gap deficit, TTC deficit, and lane mismatch. The numerical weights are fixed for the reported runs and summarized in the supplementary material.

\paragraph{Authority as a threshold parameter.} Authority $\rho$ selects one of three replan thresholds:
\begin{equation}
\tau_{\rho} =
\begin{cases}
\tau_{\mathrm{low}} & \text{if } \rho = \mathtt{low}\\
\tau_{\mathrm{med}} & \text{if } \rho = \mathtt{med}\\
\tau_{\mathrm{high}} & \text{if } \rho = \mathtt{high}
\end{cases}
\label{eq:authority_threshold}
\end{equation}
Authority does not affect hard safety: an abort-atom satisfaction or a runtime safety violation triggers override regardless of $\rho$. Authority controls only the drift-tolerance band within which the forecast continues to be reused; the configured threshold values are reported in the supplementary material.

\paragraph{Invalidation predicate.} A forecast $F$ is invalidated at step $t$ iff
\begin{equation}
\begin{split}
\mathrm{invalid}(F, S_t) :=\;& t > t_{\mathrm{issue}} + h \;\vee\; d(F, S_t) > \tau_{\rho} \\
&\vee\; \exists v \in V: \neg v(S_t) \;\vee\; \exists \alpha \in A: \alpha(S_t).
\end{split}
\label{eq:invalid_predicate}
\end{equation}
The four disjuncts correspond to expiry, drift breach, validity-atom violation, and abort-atom firing. When $\mathrm{invalid}(F, S_t)$ is true the runtime emits the analytical fallback and triggers a refresh request to the strategic layer; otherwise the runtime may preserve the forecast, soften it after completion, or override it under hard safety pressure. The shared-agent variant collapses strategic and runtime decision provenance into one model call; the dual-agent variant separates them.

\paragraph{Safety-reuse invariant.} The runtime never executes a buffered action against a state in which any abort atom fires, any validity atom is violated, the drift threshold is exceeded, or the horizon has expired. In these cases (\ref{eq:invalid_predicate}) holds and the analytical fallback action is emitted within one step. The buffered LLM commits a structured world-line choice once, and the runtime reuses that choice only while live evidence says it still applies; this safety-reuse invariant is the basis for the $H{\approx}4$~s amortization result in \S\ref{sec:results}.

\paragraph{Scope and variants.} The canonical method in this paper is the dual-agent buffered runtime, where one LLM call issues the strategic forecast and a separate runtime call consumes it under the validity, drift, abort, and safety checks above. The shared-agent runtime remains a simplified implementation variant of the same forecast-buffer contract, but it is not analyzed as a separate experimental condition in the current protocol. Deterministic replay is retained only as a contract diagnostic: it tests forecast execution and fallback predicates without live runtime LLM arbitration. The reported variants are evaluated as mechanism probes, not as separate claims of safety superiority over reactive LLM driving.

\section{Experiments}

\subsection{Protocol and Episode Definition}
\label{sec:env_setup}

We evaluate Steins;Gate Drive on closed-loop highway driving using the \texttt{highway-env} simulator with IDM-controlled traffic~\cite{treiber2000idm}. Environment parameters and the fixed seed bank are reported in Table~\ref{tab:env_setup}; values follow \texttt{highway-env} defaults for the \texttt{highway-fast-v0} configuration except where noted. The canonical evidence in this draft is a reduced matched-seed protocol: normal simulator rollouts with $n{=}10$ seeds and 20 decision steps per reported cell. Each reported condition uses the same 10 seeds, so comparisons are paired by seed with the matched episode as the unit of analysis. No curated case-set or scenario-suite evidence is used; scenario-labeled interaction-risk evaluation remains future confirmatory work.

\begin{table}[t]
\centering
\caption{Environment and episode parameters. Values follow \texttt{highway-env} defaults for \texttt{highway-fast-v0} except where labeled ours. Seeds are a fixed sample drawn from the default seed bank without scenario stratification; the matched-seed design is reported at the canonical $n{=}10$, 20-step scale. Note: reported runs apply the realistic-highway speed profile (target band $0$--$130$~km/h, reward band $80$--$130$~km/h); the discrete $\{0,5,\ldots,30\}$~m/s targets above are the simulator's primitive action proxies.}
\label{tab:env_setup}
{\footnotesize
\setlength{\tabcolsep}{4pt}
\renewcommand{\arraystretch}{1.10}
\begin{tabular}{@{}lll@{}}
\toprule
Group & Parameter & Value \\
\midrule
Environment & Simulator & \texttt{highway-env} \\
Environment & Configuration & \texttt{highway-fast-v0} \\
Environment & Lanes & 3 \\
Environment & Surrounding traffic & 20 IDM vehicles \\
Environment & Observation & Kinematics (\emph{ours}) \\
Environment & Action space & \texttt{DiscreteMetaAction}, $|\mathcal{A}|{=}5$ \\
Environment & Target speeds & $\{0,5,10,15,20,25,30\}$~m/s \\
Episode & Length & 20 decision steps (\emph{ours}) \\
Episode & Simulation frequency & 5~Hz \\
Episode & Policy frequency & 1~Hz \\
Sampling & Seed-bank source & \texttt{steinsgate\_v1} default \\
Sampling & Seed count & $n=10$ per reported cell \\
Sampling & Stratification & none (normal-run scope) \\
\bottomrule
\end{tabular}
}
\end{table}

\subsection{Experimental Conditions}
\label{sec:exp_conditions}

The protocol compares three implemented conditions, each tied to a mechanism question that the current evidence can support. We refer to conditions by scientific labels $M_1$--$M_3$ and retain implementation codenames in parentheses for reproducibility; \S\ref{sec:results} reports outcomes against this diagnostic design.

\textbf{$M_1$ Reactive (no buffer).} A single-call LLM selects the next action from the live scene without buffered foresight (codename \texttt{reactive}). This condition exposes direct-driver response latency and behavior when no forecast buffer is available.

\textbf{$M_2$ Dual-Agent Buffered Runtime (canonical method).} Strategic and runtime decisions are issued by separate LLM calls; the runtime consumes the active forecast under the predicate of (\ref{eq:invalid_predicate}) (codename \texttt{steinsgate\_dual\_agent\_runtime}). This is the canonical Steins;Gate Drive architecture and the primary condition for studying effective lag, forecast reuse, validity checks, and override behavior.

\textbf{$M_3$ Deterministic Replay (contract diagnostic).} Buffered forecasts are committed and re-executed deterministically without runtime LLM arbitration (codename \texttt{steinsgate\_planner\_deterministic}). $M_3$ is not a deployable LLM driver; it isolates forecast reuse, validity predicates, abort predicates, and fallback execution from language-level runtime judgment.

Comparisons across conditions are paired by seed, but we report descriptive rates and means rather than treating the reduced-scale study as a definitive statistical safety evaluation. Lower-score selection is reported as a selection-behavior signal; scenario-stratified safety claims require the confirmatory extension identified in \S\ref{sec:limitations}.

\subsection{Metrics and Evidence Provenance}
\label{sec:metrics}

The headline outcome metric is the per-episode no-collision indicator; secondary metrics (mean speed, accel/decel flap rate, TTC and headway tail risk, strict parse rate, lower-score selection rate, decision latency, effective residual lag, and prompt/completion token cost per decision) characterize behavior, safety tail, interface health, and runtime cost, and are reported as descriptive diagnostics rather than alternative headline claims. For latency-amortization results, we report effective residual lag rather than raw wall-clock latency:
\begin{equation}
L_{\mathrm{eff}} =
\begin{cases}
\bar{L}_{\mathrm{dec}}, & \text{reactive direct driver},\\
\bar{L}_{\mathrm{sel}}+\bar{L}_{\mathrm{dec}}-H, & \text{buffered forecast runtime},
\end{cases}
\label{eq:effective_lag}
\end{equation}
where $\bar{L}_{\mathrm{sel}}$ is the mean strategic-selector latency, $\bar{L}_{\mathrm{dec}}$ is the mean runtime decision latency, and $H$ is the forecast-reuse horizon. The default is $H{=}3$~s; the horizon sweep uses its swept value. Negative $L_{\mathrm{eff}}$ means the committed forecast covers more future time than the measured LLM call cost, not that the physical controller has negative latency. Each matched episode produces three evidence categories archived alongside results: (E1) the closed-loop outcome trace; (E2) the forecast ledger with per-strategic-call branch set, selection, and validity record; and (E3) the prompt-contract counters with parse rate and fallback events. Some buffered cells have telemetry reconstructed from per-call logs rather than aggregate metric files; reconstructed cells preserve outcome and ledger fields but not token or decision-latency accounting, and Table~\ref{tab:main_results} reports only cells with complete instrumentation. The forecast-buffered amortization claim is evaluated through $L_{\mathrm{eff}}$ and the horizon sweep; the selection-behavior claim is evaluated through lower-score selection rate.

\section{Results}
\label{sec:results}

Table~\ref{tab:horizon_resolved} previews the main latency-amortization result: forecast reuse moves GPT-5.4 mini from positive to near-zero effective lag while no-collision remains saturated. The rest of this section is organized around deployment conditions and mechanism knobs rather than internal run phases. We first compare direct reactive LLM control, deterministic forecast replay, and the canonical dual-agent runtime, then analyze how role exposure, visible budget, horizon, and drift threshold affect the forecast-buffer interface.

All main rows use the reduced realistic-highway \texttt{n10}/20-step protocol with the same matched seed bank, realistic speed profile, strict parseable response contract, and safety-aligned validity/abort evaluation. Partially reconstructed longer-run rows are excluded because they do not preserve complete token and latency accounting; thus Table~\ref{tab:main_results} reports only clean, fully instrumented cells.

\subsection{Cross-Model Screening and Deployability Diagnostics}

Table~\ref{tab:main_results} compares the three conditions on matched \texttt{n10}/20-step cells. The dual-agent runtime preserves the no-collision floor while deterministic replay isolates how much behavior is carried by the typed contract alone: validity, abort, and fallback. The table does not establish a safety-superiority claim over the reactive baseline, because the normal-highway protocol is collision-saturated for several direct and buffered rows.

\begin{table*}[!t]
\centering
\caption{Main \texttt{n10}/20-step screening results. Red rows are negative evidence for deployability; the clean Claude Haiku 4.5 dual-agent row is pending and omitted. Bold marks the best comparable deployable value and underlining marks the second-best value within a block.}
\label{tab:main_results}
{\scriptsize
\setlength{\tabcolsep}{3.1pt}
\renewcommand{\arraystretch}{1.12}
\begin{tabular*}{\textwidth}{@{\extracolsep{\fill}}>{\raggedright\arraybackslash}p{0.055\textwidth}
>{\raggedright\arraybackslash}p{0.125\textwidth}
cccccccc@{}}
\toprule
Mode & Model &
\shortstack{No-coll.\\\%$\uparrow$} &
\shortstack{Comp.\\\%$\uparrow$} &
\shortstack{Speed\\km/h$\uparrow$} &
\shortstack{Flap\\\%$\downarrow$} &
\shortstack{TTC\\\%$\downarrow$} &
\shortstack{Eff.\\lag s$\downarrow$} &
\shortstack{Tok. k\\P/C$\downarrow$} &
\shortstack{Low\\\%} \\
\midrule
\multicolumn{10}{@{}l}{\emph{Direct LLM driving baseline}} \\
\rowcolor{red!10} Reactive & DeepSeek-v4-flash & 100.0 & 100.0 & 81.0 & 21.0 & 0.5 & 20.31 & 1.34/1.37 & n/a \\
Reactive & Claude Haiku 4.5 & 100.0 & 100.0 & \second{72.0} & 38.0 & \best{0.0} & 2.05 & 1.44/0.11 & n/a \\
Reactive & GPT-5.4 mini & 100.0 & 100.0 & \best{75.3} & \second{21.5} & 1.0 & \best{0.97} & 1.26/0.03 & n/a \\
Reactive & Qwen3-235B-A22B & 100.0 & 100.0 & 67.6 & \best{17.0} & \best{0.0} & \second{1.86} & 1.37/0.05 & n/a \\
\rowcolor{red!10} Reactive & Llama-3.2-3B & 100.0 & 100.0 & 9.5 & 0.0 & 0.0 & 0.28 & 1.19/0.01 & n/a \\
\midrule
\multicolumn{10}{@{}l}{\emph{Deterministic forecast replay negative control}} \\
\rowcolor{red!10} Det. & DeepSeek-v4-flash & 100.0 & 100.0 & 80.9 & 16.5 & 0.0 & 17.46 & 0.16/0.13 & 8.7 \\
Det. & Claude Haiku 4.5 & 100.0 & 100.0 & \second{80.0} & \second{16.0} & \best{0.0} & \second{0.26} & 0.18/0.01 & 3.9 \\
Det. & GPT-5.4 mini & 100.0 & 100.0 & 75.8 & \best{12.0} & \best{0.0} & \best{-1.22} & 0.16/0.00 & 19.1 \\
Det. & Qwen3-235B-A22B & 90.0 & 94.5 & \best{81.9} & 24.0 & 3.3 & 2.85 & 0.18/0.01 & 30.2 \\
\rowcolor{red!10} Det. & Llama-3.2-3B & 0.0 & 0.0 & 0.0 & 0.0 & 0.0 & -0.80 & 1.22/0.01 & n/a \\
\midrule
\multicolumn{10}{@{}l}{\emph{Dual-agent Steins;Gate Drive screening}} \\
\rowcolor{red!10} Dual & DeepSeek-v4-flash & 100.0 & 100.0 & 80.2 & 19.5 & 0.0 & 51.16 & 1.37/1.18 & 18.7 \\
Dual & GPT-5.4 mini & 100.0 & 100.0 & \best{74.1} & \best{9.0} & 0.5 & \best{0.40} & 1.29/0.03 & 17.0 \\
Dual & Qwen3-235B-A22B & 100.0 & 100.0 & \second{72.8} & 39.0 & \best{0.0} & 5.64 & 1.42/0.04 & 37.8 \\
\rowcolor{red!10} Dual & Llama-3.2-3B & 100.0 & 100.0 & 22.6 & 68.0 & 0.0 & -1.03 & 1.13/0.01 & 0.0 \\
\bottomrule
\end{tabular*}
\vspace{1pt}
\begin{minipage}{\textwidth}
\scriptsize
\emph{Note.} Values are seed means over matched \texttt{n10}/20-step cells; percentage cells omit the percent sign. For buffered rows, Eff. lag $=$ Sel.~s $+$ Dec.~s $-H$ with $H{=}3$~s; for reactive rows it is the direct decision latency because no forecast horizon is reused. Tok. k P/C reports prompt/completion kilotokens per decision. Low \% reports the rate at which the selector chooses a branch below the analytical top score; \texttt{n/a} denotes a non-applicable field. Uncertainty intervals and paired tests for the main rows are summarized in the supplementary material. Red rows follow explicit table diagnostics: Eff. lag $>10$~s, mean speed $<30$~km/h, flap $>50\%$, or zero-completion deterministic failure.
\end{minipage}
}
\end{table*}

Two patterns matter for the rest of the paper. First, reactive rows can also complete this normal-highway protocol, so the contribution is not a higher raw no-collision rate. Second, the same table separates deployability failures that a single safety number would hide: DeepSeek-v4-flash is collision-free but latency-prohibitive, Llama-3.2-3B is collision-free in dual mode but slow and oscillatory, and GPT-5.4 mini is the only clean dual-agent row that combines completion with near-zero effective lag. Deterministic replay is retained only as a contract diagnostic because it removes live runtime LLM arbitration while preserving forecast validity, abort, and fallback predicates.

\subsection{Mechanism Sensitivity Sweeps}

Table~\ref{tab:sweep_sensitivity} characterizes how the architecture responds to its main mechanism knobs under one fixed model: GPT-5.4 mini on the realistic-highway \texttt{n10}/20-step protocol. The sweeps vary role exposure, selector-visible worldline budget, forecast horizon, and drift threshold while holding the runtime contract and seed bank fixed.

\begin{table*}[!t]
\centering
\caption{Mechanism sensitivity sweeps for Steins;Gate Drive using GPT-5.4 mini on matched normal \texttt{highway-fast-v0} \texttt{n10}/20-step runs. Each block varies one mechanism knob and reports safety, speed, effective lag, and lower-score selection.}
\label{tab:sweep_sensitivity}
{\scriptsize
\setlength{\tabcolsep}{3.6pt}
\renewcommand{\arraystretch}{1.10}
\begin{tabular*}{\textwidth}{@{\extracolsep{\fill}} l c c c c c >{\itshape}c @{}}
\toprule
Method / Metric & \multicolumn{5}{c}{Setting (left $\rightarrow$ right)} & Trend \\
\cmidrule(lr){2-6}
\multicolumn{7}{@{}l}{\emph{Role-set/balance sweep: $\alpha$, $\alpha{+}\beta$, $\alpha{+}\gamma$, $\alpha{+}\beta{+}\gamma$ balanced, $\alpha{+}\beta{+}\gamma$ natural}} \\
Dual / No-coll. \%$\uparrow$ & 100.0 & 100.0 & 100.0 & 100.0 & 100.0 & -- \\
Dual / Speed km/h$\uparrow$ & 75.4 & 73.4 & 67.2 & 72.8 & 76.3 & -- \\
Dual / Eff. lag s$\downarrow$ & +0.57 & +0.52 & \best{+0.24} & \second{+0.29} & +1.21 & -- \\
Dual / Low \% & 28.3 & 74.7 & 70.9 & 87.0 & 10.4 & -- \\
\midrule
\multicolumn{7}{@{}l}{\emph{Worldline-budget sweep: selector-visible budget $k \in \{1,\ 2,\ 3,\ 6,\ 8\}$}} \\
Dual / No-coll. \%$\uparrow$ & 100.0 & 100.0 & 100.0 & 100.0 & 100.0 & +0.0 \\
Dual / Speed km/h$\uparrow$ & 75.1 & 73.2 & 74.0 & 74.2 & 70.9 & -4.2 \\
Dual / Eff. lag s$\downarrow$ & \best{+0.78} & +0.97 & \second{+0.93} & +1.19 & +1.27 & +0.48 \\
Dual / Low \% & 0.0 & 91.3 & 86.1 & 76.5 & 82.6 & +82.6 \\
\midrule
\multicolumn{7}{@{}l}{\emph{Horizon sweep: coupled foresight horizon $H \in \{1,\ 2,\ 4,\ 8,\ 12\}$ seconds/steps}} \\
Dual / No-coll. \%$\uparrow$ & 100.0 & 100.0 & 100.0 & 100.0 & 100.0 & +0.0 \\
Dual / Speed km/h$\uparrow$ & \best{73.4} & \second{70.2} & 68.7 & 67.1 & 50.0 & -23.4 \\
Dual / Eff. lag s$\downarrow$ & +3.07 & +1.97 & -0.01 & \second{-3.98} & \best{-7.79} & -10.86 \\
Dual / Low \% & 93.8 & 93.2 & 87.5 & 91.1 & 91.9 & -1.9 \\
\midrule
\multicolumn{7}{@{}l}{\emph{Drift-threshold sweep: $\tau_{\rho} \in \{0.20,\ 0.35,\ 0.50,\ 0.65,\ 1.00\}$; $\tau{=}1.00$ disables drift-based invalidation}} \\
Dual / No-coll. \%$\uparrow$ & 100.0 & 100.0 & 100.0 & 100.0 & 100.0 & +0.0 \\
Dual / Speed km/h$\uparrow$ & 73.3 & 70.0 & 72.5 & 69.8 & 71.6 & -1.7 \\
Dual / Eff. lag s$\downarrow$ & \best{+1.25}$^{\dagger}$ & +1.56 & +1.28 & +1.51 & \best{+1.25}$^{\dagger}$ & 0.00 \\
Dual / Low \% & 84.3 & 87.4 & 83.4 & 87.8 & 87.4 & +3.0 \\
\bottomrule
\end{tabular*}
}
\end{table*}

\paragraph{Role balance activates stress evidence.} Lower-score selection rises from $28.3\%$ under alpha-only branches to $87.0\%$ under the balanced full $\alpha{+}\beta{+}\gamma$ taxonomy, while no-collision remains unchanged. The natural full-taxonomy control is revealing: with the same role set but without balanced role pressure, lower-score selection falls to $10.4\%$, mean speed rises to $76.3$~km/h, and TTC danger appears at $1.5\%$. Beta and gamma evidence is not used simply because it is visible; balanced presentation is what turns stress branches into active decision evidence.

\paragraph{Visible budget unlocks semantic divergence early.} The budget sweep shows a step from 0\% lower-score selection at $k{=}1$ (no choice possible) to $91.3\%$ at $k{=}2$, then remains high through $k{=}8$ ($82.6\%$). The first additional branch is what unlocks semantic divergence; beyond that, larger visible budgets mostly trade efficiency and latency rather than changing the qualitative selection regime.

\paragraph{Horizon trades latency for conservative execution.} The horizon block of Table~\ref{tab:sweep_sensitivity} and the horizon-resolved Table~\ref{tab:horizon_resolved} both show effective lag dropping from $+3.07$~s at $H{=}1$ to $-7.79$~s at $H{=}12$, while no-collision remains at $100\%$. Speed declines slowly through $H{=}8$ and then drops to $50.0$~km/h at $H{=}12$, indicating that buffer staleness manifests as conservative driving rather than unsafe driving on this protocol.

\paragraph{Safety saturation bounds interpretation.} All 20 \emph{No-coll.} cells across the four sweeps report 100\%. Whether we vary the role set, role-balance prompt, worldline budget $k$, horizon $H$, or drift threshold $\tau_{\rho}$, no swept configuration produces a collision on the realistic-highway \texttt{n10}/20-step protocol. Mechanism evidence is reported through speed, selection behavior, TTC danger, and effective latency rather than through collision-rate differences.

\paragraph{Drift is a refresh-frequency knob.} The five-cell drift-threshold sweep shows that $\tau_{\rho}{=}1.00$, which disables drift-based invalidation entirely, preserves $100\%$ no-collision. Speed varies non-monotonically across thresholds ($73.3$, $70.0$, $72.5$, $69.8$, and $71.6$~km/h), and effective lag is nearly unchanged between the ultra-conservative and disabled endpoints ($1.25$~s in both cases). On this protocol, the abort-atom and validity-atom predicates of the typed forecast contract carry the main safety load, rather than the drift score.

\paragraph{Deterministic replay is a contract diagnostic.} Deterministic-replay viability in Table~\ref{tab:main_results} is consistent with the drift-sweep reading because deterministic mode enforces the same atom predicates without a runtime LLM in the loop. This should be read as a diagnostic for the typed forecast contract, not as evidence that deterministic replay is a deployable controller. Taken together, the four swept knobs separate into selection-behavior knobs, an execution-behavior knob, and a refresh-frequency knob: role exposure and visible budget govern what the LLM picks, horizon governs how long the runtime executes the picked forecast, and drift threshold governs how often the system refreshes the strategic buffer. No swept knob affects no-collision on this saturated protocol, so the safety floor is carried by mechanisms that do not vary across these sweeps---specifically, validity atoms, abort atoms, and analytical fallback.

\subsection{Horizon-Resolved Latency Amortization}

The horizon-resolved Table~\ref{tab:horizon_resolved} expands the horizon block of Table~\ref{tab:sweep_sensitivity}. Effective control-loop lag drops from $+3.07$~s at $H{=}1$ (forecast refreshed every step, no amortization) to $-7.79$~s at $H{=}12$ (forecast covers $60\%$ of the episode), crossing zero between $H{=}2$ and $H{=}4$. This knee point at $H{\approx}4$~s is the operational meaning of ``foresight helps'': below it, the architecture pays raw strategic and runtime call costs without amortization; above it, the buffered horizon covers more time than the call sequence requires, and the LLM's contribution shifts from a per-step controller to a structured-future arbitrator whose cost is spread over multiple control steps.

The speed trade-off matters as much as the knee point itself. Speed declines monotonically from $73.4$~km/h at $H{=}1$ to $50.0$~km/h at $H{=}12$, while no-collision rate remains at $100\%$ across all five horizon settings. Under long horizons, the runtime's atom predicates and analytical fallback hold safety constant while outdated commitments translate into conservative driving rather than crashes.

\subsection{Role-Set Mechanism Audit}

Table~\ref{tab:role_resolved} gives the complementary role-set view. The controlled GPT-5.4 mini role-set sweep asks whether beta and gamma world lines can actively enter the selected forecast when they are exposed to the selector, and how this changes under balanced versus natural prompt presentation. The natural column is fastest (76.3~km/h), but it pays the worst effective lag ($+1.21$~s), the highest flap rate (17.5\%), the only visible TTC danger (1.5\%), and exercises the stress-role taxonomy only $3.0\%$ of the time. The balanced full taxonomy exercises the architecture's mechanism by lifting stress-role selection to $87.0\%$, but it does so at a small speed and latency cost.

\begin{table*}[t]
\centering
\caption{Role-set mechanism audit for the dual-agent Steins;Gate Drive selector. Values are from the GPT-5.4 mini realistic-highway \texttt{n10}/20-step sweep. The first four columns are balanced presentations that expose at least one branch of each enabled role to the selector; the fifth column (nat.) keeps the implementation's unbalanced role availability. Bold marks the best comparable value and underlining the second-best for directed metric rows.}
\label{tab:role_resolved}
{\scriptsize
\setlength{\tabcolsep}{3.7pt}
\renewcommand{\arraystretch}{1.10}
\begin{tabular*}{\textwidth}{@{\extracolsep{\fill}}lccccc@{}}
\toprule
Metric & $\alpha$ & $\alpha{+}\beta$ & $\alpha{+}\gamma$ & $\alpha{+}\beta{+}\gamma$ bal. & $\alpha{+}\beta{+}\gamma$ nat. \\
\midrule
No-collision rate \%$\uparrow$ & 100.0 & 100.0 & 100.0 & 100.0 & 100.0 \\
Speed km/h$\uparrow$ & \second{75.4} & 73.4 & 67.2 & 72.8 & \best{76.3} \\
Flap \%$\downarrow$ & 9.5 & 11.0 & \best{5.5} & \second{9.5} & 17.5 \\
TTC danger \%$\downarrow$ & \best{0.0} & \best{0.0} & \best{0.0} & \best{0.0} & 1.5 \\
Eff. lag s$\downarrow$ & +0.57 & +0.52 & \best{+0.24} & \second{+0.29} & +1.21 \\
Low-score sel. \% & 28.3 & 74.7 & 70.9 & 87.0 & 10.4 \\
Sel. $\alpha$ \% & 100.0 & 28.8 & 33.0 & 13.0 & 97.0 \\
Sel. $\beta$ \% & 0.0 & 71.2 & 0.0 & 58.7 & 2.2 \\
Sel. $\gamma$ \% & 0.0 & 0.0 & 67.0 & 28.3 & 0.9 \\
Stress-role sel. $\beta{+}\gamma$ \%$\uparrow$ & 0.0 & \second{71.2} & 67.0 & \best{87.0} & 3.0 \\
Same-action non-$\alpha$ sel. \%$\uparrow$ & n/a & 79.5 & \second{84.9} & \best{90.1} & 5.8 \\
$\alpha$ default despite non-$\alpha$ \%$\downarrow$ & n/a & 20.5 & \second{15.1} & \best{9.9} & 94.2 \\
\bottomrule
\end{tabular*}
\vspace{1pt}
\begin{minipage}{\textwidth}
\scriptsize
\emph{Note.} The table is a controlled mechanism audit, not a cross-model performance comparison. Balanced columns rotate a per-step \texttt{role\_priority} and instruct the selector to compare $\alpha$, $\beta$, and $\gamma$ as equal evidence hypotheses; the natural column omits that instruction and retains analytical pruning's organic role mix. \emph{Same-action non-$\alpha$} asks whether the LLM chooses a $\beta/\gamma$ stress branch when one exists for the same primitive action; the final row reports the complementary $\alpha$-default rate.
\end{minipage}
}
\end{table*}

The purpose of reporting lower-score rates in Tables~\ref{tab:main_results}, \ref{tab:sweep_sensitivity}, and~\ref{tab:role_resolved} is not to claim that lower analytical score is automatically better. It asks whether the LLM is merely copying the analytical top branch or sometimes selecting a different semantic future. In the current evidence, this remains diagnostic only: lower-score selections motivate replay or offline scoring of the unchosen top branches, but they do not by themselves prove counterfactual safety or comfort improvement. Establishing such a claim requires counterfactual replay or scenario-stratified seeds that compare the selected branch against the unchosen analytical-top alternative under the same scene state.

In sum, the cross-model rows establish deployability boundaries for the forecast-buffer interface; the horizon sweep establishes latency amortization around $H{\approx}4$~s; and the role-set sweep establishes that beta/gamma stress evidence becomes active only when the selector is prompted to compare roles as equal evidence hypotheses. These three results instantiate the paper's claims about bounded world-line generation, typed forecast contracts, and latency-decoupled runtime arbitration without claiming safety superiority over reactive LLM driving.

\section{Discussion}

\subsection{Design Implications}

The results support three design lessons. First, buffering helps only when the reuse horizon exceeds the combined call cost. Second, beta and gamma futures matter only when exposed as comparable alternatives, not as incidental branches dominated by alpha rollouts. Third, validity atoms, abort atoms, fallback actions, and runtime authority carry safety; collision-free rows may still be undeployable if they are slow, oscillatory, or fallback-dependent.

\subsection{Positioning}

Steins;Gate Drive sits between direct LLM driving, lower-frequency LLM planning, learned world models, and classical safety interfaces. It separates semantic future choice from runtime execution, makes slow commitments revocable through atom predicates, and avoids learning a visual simulator or latent planner. Its closest classical analogy is MPC/RSS discipline: future-conditioned planning is useful, but safety checks must remain independent of the planner's internal reasoning.

The same contract pattern can apply when a slow semantic model makes an early choice supervised by a faster runtime. Robotics, drones, and industrial control would need domain-specific state encodings and safety predicates, but the reusable idea is the same: commit a structured forecast, then revoke it when live conditions violate the contract.

\section{Limitations and Future Work}
\label{sec:limitations}

\subsection{Threats to Validity}

The evidence remains scoped to normal \texttt{highway-env} runs with simplified IDM traffic, no perception uncertainty, and a realistic-highway speed profile. The main evidence uses completed $n{=}10$/20-step cells, many with $100\%$ no-collision, so the paper does not claim safety superiority over reactive LLM driving or real-world transfer. GPT-5.4 mini provides the sweep evidence; other models show different deployability and lower-score-selection behavior.

\subsection{Future Work}

Future work should regenerate any reconstructed cells with complete artifacts, add scenario labels, replay or score unchosen analytical-top alternatives, extend paired tests, and scale to $n{\geq}50$ matched seeds. Adaptive branch generation, risk-sensitive scoring, cooperative foresight, and learned-world-model integration require driving-environment validation.

\paragraph{Data and supplementary material.}

The companion supplement reports uncertainty intervals, prompt contracts, branch-generation hyperparameters, pseudocode, reproducibility details, failure cases, and artifact summaries.

\section{Conclusion}

In completed $n{=}10$/20-step cells, the GPT-5.4 mini dual-agent runtime reaches near-zero effective lag around $H{\approx}4$~s while fallback and atom predicates preserve the runtime safety floor. Steins;Gate Drive makes slow LLM reasoning executable by converting world-line choice into a revocable \texttt{StrategicForecast} over branch choice, validity, abort conditions, fallback behavior, and runtime authority. The claim is narrow but actionable: forecast-buffered LLM driving can amortize slow reasoning when the horizon covers call cost, while live arbitration remains responsible for immediate safety.

\bibliographystyle{IEEEtran}
\bibliography{steinsgatedrive_refs}

\begin{IEEEbiography}[{\includegraphics[width=1in,height=1.25in,clip,keepaspectratio]{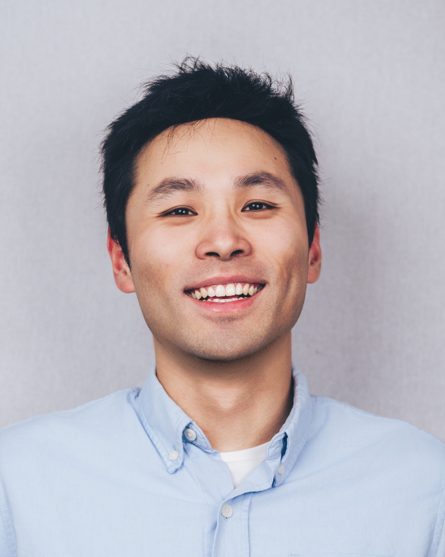}}]{Anjie Qiu}
received the B.S. degrees from Fuzhou University, Fujian, China, and the Technical University of Kaiserslautern, Germany, in 2018, and the M.S. degree from the University of Kaiserslautern, Germany, in 2019. He is pursuing the Ph.D. degree with the Institute for Wireless Communication and Navigation, RPTU University Kaiserslautern-Landau, Germany. His research interests include LLM/VLA reasoning, latency-aware autonomous driving, simulation-backed evaluation, and typed agent interfaces.
\end{IEEEbiography}

\begin{IEEEbiography}[{\includegraphics[width=1in,height=1.25in,clip,keepaspectratio]{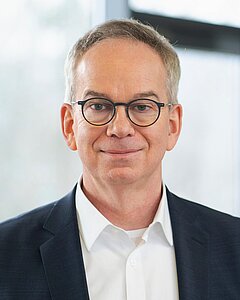}}]{Hans D. Schotten}
studied electrical engineering at RWTH Aachen University, Germany, from 1984 to 1990, and received the Dr.-Ing. degree there in 1997. He is Professor of Radio Communication and Navigation at RPTU University Kaiserslautern-Landau and Scientific Director of the Intelligent Networks Research Department at DFKI, Kaiserslautern, Germany. He previously held research and standards leadership roles at Ericsson Corporate Research and Qualcomm. His research interests include 5G/6G communications, industrial wireless automation, resilient networks, and automotive and railway communication systems.
\end{IEEEbiography}

\end{document}